\documentclass{article}

\usepackage{color}

\usepackage{epsf}

\usepackage{psfrag}
\usepackage{graphicx}
\usepackage{amssymb}
\usepackage{amsbsy}
\usepackage{epsfig}
\usepackage{ifthen}
\usepackage{eucal}
\usepackage{theorem}

\newcommand{\commentout}[1]{}

\newcommand{\url}[1]{{\tt\small #1}}

\newcommand{\Ncal}{\mathcal{N}} 
\newcommand{\Xcal}{\mathcal{X}} 
\newcommand{\Ycal}{\mathcal{Y}} 

\newcommand{\fig}[1]{Figure~\protect\ref{#1}}

\renewcommand{\sec}[1]{Section~\protect\ref{#1}}

\newcommand{\eq}[1]{(\protect\ref{#1})}

\newcommand{\independent}{\perp\mkern-11mu\perp}

  \mathchardef\ordinarycolon\mathcode`\:
  \mathcode`\:=\string"8000
  \begingroup \catcode`\:=\active
    \gdef:{\mathrel{\mathop\ordinarycolon}}
  \endgroup

\newtheorem{Ex}{Example}
\newtheorem{Def}{Definition}

\usepackage{caption}

\begin{document}

\def\short{1}
\def\long{}
\def\version{\short}        
\def\iflong{\ifnum\version=\long}
\def\ifshort{\ifnum\version=\short}

\begin{centering}
{\Large\bf Robust Learning via Cause-Effect Models}

\bigskip

Bernhard Sch\"olkopf, Dominik Janzing, Jonas Peters \& Kun Zhang

\smallskip
Max Planck Institute for Intelligent Systems\\
Spemannstr.\ 38\\
T\"ubingen, Germany

\smallskip
\{first.last@tuebingen.mpg.de\}

\end{centering}

\begin{abstract}
We consider the problem of function estimation in the case where the data distribution may shift between training and test time, and additional information about it may be available at test time. This relates to popular scenarios such as covariate shift, concept drift, transfer learning and semi-supervised learning. 

This working paper discusses how these tasks could be tackled depending on the kind of changes of the distributions. It argues that knowledge of an underlying causal direction can facilitate several of these tasks.

\end{abstract}

\section{Introduction}

By and large, statistical machine learning exploits statistical associations or dependences between variables to make predictions about certain variables. This is a very powerful concept especially in situations where we have sizable training sets, but no detailed model of the underlying data generating process. This process is usually modelled as an unknown probability distribution, and machine learning excels whenever this distribution does not change. Most of the theoretical analysis assumes that the data are i.i.d.\ (independent and identically distributed) or at least exchangeable. 

On the other hand, practical problems often do not have these favorable properties, forcing us to leave the comfort zone of i.i.d.\ data. Sometimes distributions shift over time, sometimes we might want to combine data recorded under different conditions or from different but related regularities. Researchers have developed a number of modifications of statistical learning methods to handle various scenarios of changing distributions, for an overview, see \cite{SugKaw12}.

The present paper attempts to study these problems from the point of view of causal learning. 
As some other recent work in the field \cite{LemeireD,JanSch10}, it will build on the assumption that in causal structures, the distribution of the cause and the mechanism
relating cause and effect tend to be independent\footnote{In the mentioned references, independence is meant in the sense of algorithmic independence, but other notions of independence can also make sense.}. For instance, in the problem of predicting splicing patterns from genomic sequences, the basic splicing mechanism (driven by the ribosome) may be assumed stable betwen different species \cite{SchWidSchRat09}, even though the genomic sequences and their statistical properties might differ in several respects. This is important information constraining causal models, and it can also be useful for robust predictive models as we try to show in the present paper. Intuitively, if we learn a causal model of splicing, we could hope to be more robust with respect to changes of the input statistics, and we may be able to combine data collected from different species to get a more accurate statistical model of the splicing mechanism.

Causal graphical models as pioneered by \cite{Pearl00,SpiGlySch93} are usually thought of as joint probability distributions over a set of variables $X_1,\dots,X_n$, along with directed graphs (for simplicity, we assume acyclicity) with vertices $X_i$ and arrows indicating direct causal influences. The \emph{causal Markov assumption} \cite{Pearl00} states that each vertex $X_i$ is independent of its non-descendants in the graph, given its parents. Here, independence is usually meant in a statistical sense, although alternative views have been developed, e.g., using algorithmic independence \cite{JanSch10}. 

Crucially, the causal Markov assumption links the semantics of causality to something that has empirically measurable consequences (e.g., conditional statistical independence). Given a sufficient set of observations from a joint distribution, it allows us to test conditional independence statements and thus infer (subject to a genericity assumption referred to as ``faithfulness'') which causal models are consistent with an observed distribution. However, this will typically not lead us to a unique causal model, and in the case of graphs with only two variables, there are no conditional independence statements to test and we cannot do anything.

There is an alternative view of causal models, which does not start from a joint distribution. Instead, it assumes a set of jointly independent noise variables, one at each vertex, and each vertex computes a deterministic function of its noise variables and its parents. This view, referred to as a functional causal model (or nonlinear structural equation model), 
 entails a joint distribution which along with the graph satisfies the causal Markov assumption \cite{Pearl00}. Vice versa, each causal graphical model can be expressed as a functional causal model \cite[e.g.]{JanSch10}.
\footnote{As an aside, note that the functional point of view is more specific than that graphical model view \cite{Pearl00}. To see this consider $X \rightarrow Y$ and the following two functional models that lead to the same joint distribution: (1) $Y = X \mbox{~xor~} N$ with $P(N=0)=2/3, P(N=1) = 1/3$ and (2) $Y=f(X,N)=f_N(X)$ with $f_0 \equiv 0, f_1 \equiv 1, f_2 = id$ and $P(N=0)=P(N=1)=P(N=2)=1/3$. Suppose one observes the sample $(0,0)$. (1) and (2) give different answers to the counterfactual question ``What would have happened if $X$ had been one?''. The causal graph and the joint distribution does not provide sufficient information to give any answer.}

The functional point of view is rather useful in that it allows us to come up with assumptions on causal models that would be harder to conceive in a pure probabilistic view. It has recently been shown \cite{HoyJanMooPetetal09} that an assumption of nonlinear functions with additive noise renders the two variable case (and thus the multivariate case \cite{Peters2011}) identifiable, i.e.,
we can distinguish between the causal structures $X\rightarrow Y$ and $X \leftarrow Y$, given that one and only one of these two alternatives is true 
(which implicitly excludes a common cause of $X$ and $Y$). Hence, we can tackle the 
case where conditional independence tests do not provide any information. 
This opens up the possibility to identify the causal direction for input-output learning problems. The present paper assays whether this can he helpful for machine learning, and it argues that in many situations, a causal model can be more robust under distribution shifts than a pure statistical model. Perhaps somewhat suprisingly, learning problems need not always predict effect from cause, and the direction of the prediction has consequences for which tasks are easy and which tasks are hard. In the remainder of the paper, we restrict ourselves to the simplest possible case, where we have two variables only and there are no unobserved confounders.

%
%
\begin{figure}[h]
\begin{center}
\includegraphics[width=8cm]{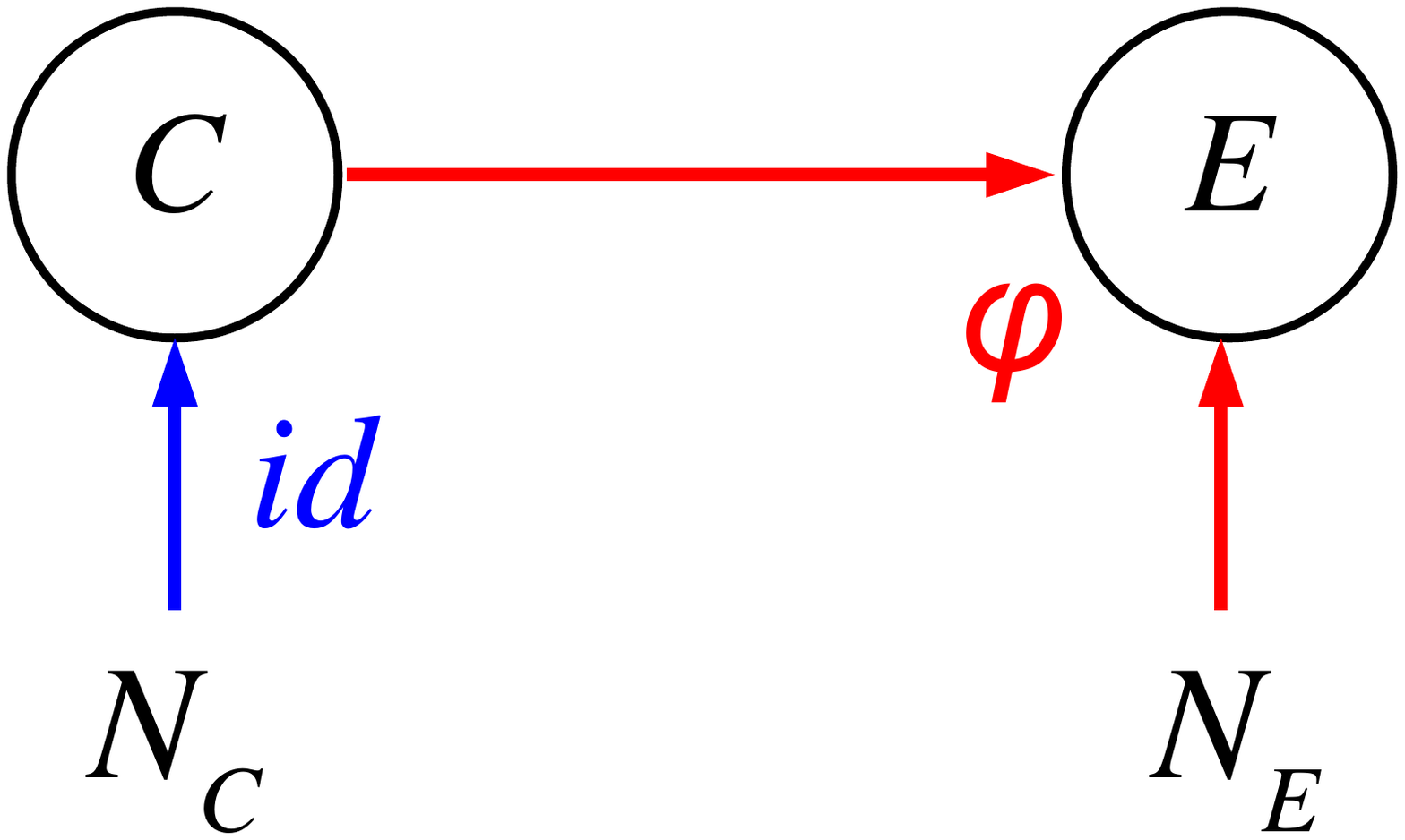}
\end{center}
\caption[font=small,labelsep=none]{\label{cause-effect-SEM}A simple functional causal model, where $C$ is the cause variable, $\varphi$ is a deterministic mechanism, and $E$ is the effect variable. $N_C$ is a noise variable influencing $C$ (without restricting generality, we can equate this with $C$), and $N_E$ influences $E$ via $E=\varphi(C,N_E)$. We assume that $N_C$ and $N_E$ are independent, in which case we may restrict our attention to the present graph (causal sufficiency). }
\end{figure}

\paragraph{Notation.}
We consider the causal structure shown in Fig.~\ref{cause-effect-SEM}, with two observables, modeled by random variables. When using the notation $C$ and $E$, the variable $C$ stands for the cause and $E$ for the effect. We denote their domains $\cal C$ and $\cal E$ and their distributions by $P(C)$ and $P(E)$ (overloading the notation $P$). When using the notation $X$ and $Y$, the variable $X$ will always be the input and $Y$ the output, from a machine learning point of view (but input and output can either be cause of effect --- more below).

For simplicity, we assume that their distributions have a joint density with respect to some product measure. We write the values of this density as $P(c,e)$ and the values of the marginal densities as $P(c)$ and $P(e)$, again keeping in mind that these three $P$ are different functions --- we can always tell from the argument which function is meant.

We identify a training set of size $l$ with a uniform mixture of Dirac measures, denoted as 
$P(C,E)$ 
and use an analogous notation for an additional data set of size $m$ (e.g., a set of test inputs). E.g., 
$P'(C)$ 
could be a set of test inputs sampled from a distribution $P'$ that need not be identical with $P$.
The following assumptions are used throughout the paper. The subsections below only mention additional assumptions that are task specific.

\paragraph{Causal sufficiency.}
We further assume that there are two independent noise variables $N_C$ and $N_E$, modeled as random variables with domains $\Ncal_C$ and $\Ncal_E$ and distributions $P(N_C)$ and $P(N_E)$.
In some places, we will use conditional densities, always implicitly assuming that they exist.

The function $\varphi$ and the noise term $P(N_E)$ jointly determine the conditional $P(E|C)$ via
\[
E=\varphi(C,N_E)\, .
\]
We think of $P(E|C)$ as the \emph{mechanism} transforming cause $C$ into effect $E$. 

\paragraph{Indepence of mechanism and input.}
We finally assume that the mechanism is ``independent'' of the distribution of the cause (i.e., independent of $C=N_C$ in Fig.~\ref{cause-effect-SEM}), in the sense that $P(E|C)$ contains no information about $P(C)$ and vice versa; in particular, if $P(E|C)$ changes at some point in time, there is no reason to believe that $P(C)$ changes at the same time.\footnote{A stronger condition, which we do not need in the present context, would be to require that $P(N_E)$, $\varphi$ and $P(C)$ be jointly ``independent.''}
This assumption has been used by \cite{LemeireD,JanSch10}. It encapsulates our belief that $\varphi$ is a mechanism of nature that does not care what we feed into it. 
The assumption introduces an important asymmetry between cause and effect, since it will usually be violated in the backward direction, i.e., the distribution of the effect $E$ will inherit properties from $\varphi$ \cite{JanSch10,DanJanMooZscSteZhaSch10}.

\paragraph{Richness of functional causal models}
It turns out that the two-variable functional causal model is so rich that it cannot be identified. The causal Markov condition is trivially satisfied both by the forward model and the backward model, and thus both graphs allow a functional model.

To understand the richness of the class intuitively, consider the simple case where the noise $N_E$ can take only a finite number of values, say $\{ 1,\dots,v\}$. This noise could affect $\varphi$ for instance as follows: there is a set of functions $\{\varphi_n \colon n=1,\dots,v\}$, and the noise randomly switches one of them on at any point, i.e.,
$$\varphi(c,n) = \varphi_n(c).$$
The functions $\varphi_n$ could implement arbitrarily different mechanisms, and it would thus be very hard to identify $\varphi$ from empirical data sampled from such a complex model.\footnote{A similar construction, with the range of the noise having the cardinality of the function class, can be  used  \cite{JanSch10} to argue that every causal graphical model can be expressed as a functional causal model.}

As an aside, recall that for acyclic causal graphs with more than two variables, the graph structure will typically imply conditional independence properties via the causal Markov condition. However, the above construction with noises randomly switching between mechanisms is still valid, and it is thus surprising that conditional independence alone does allow us to do some causal inference of practical significance, as implemented by the well known PC and FCI algorithms \cite{SpiGlySch93,Pearl00}. It should be clear that additional assumptions that prevent the noise switching construction should significantly facilitate the task of identifying causal graphs from data. Intuitively, such assumptions need to control the complexity with which the noise $N_E$ given a training set plus two unpaired sets from the two original marginals. 

\paragraph{Additive noise models.}
One such assumption is referred to as {\tt ANM}, standing for \emph{nonlinear non-Gaussian acyclic model} \cite{HoyJanMooPetetal09}. This model assumes 
$\varphi(C,N_E)=\phi(C)+N_E$ for some function $\phi$:
\begin{equation}\label{eq:ANM}
E = \phi(C) + N_E\,,
\end{equation}
and it has been shown that $\phi$ and $N_E$ can be identified in the generic case, provided that $N_E$ is assumed to have zero mean. This means that apart from some exceptions, such as the case where $\phi$ is linear and $N_E$ is Gaussian, a given joint distribution of two real-valued random variables $X$ and $Y$ can be fit by an {\tt ANM} model in at most one direction (which we then consider the causal one).

A similar statement has been shown for discrete data \cite{Peters2011b} and for the {\tt post\-non\-linear ANM} model \cite{Zhang_UAI}
$$
E = \psi(\phi(C) + N_E)\,,
$$
where $\psi$ is an invertible function.

In practice, an {\tt ANM} model can be fit by regressing the effect on the cause while enforcing that the residual noise variable is independent of the cause \cite{MooJanPetSch09}. If this is impossible, the model is incorrect (e.g., cause and effect are interchanged, the noise is not additive, or there are confounders).

{\tt ANM} plays an important role in this paper; first, because all the methods below will presuppose that we know what is cause and what is effect, and second, because we will generalize {\tt ANM} to handle the case where we have several models of the form \eq{eq:ANM} that share the same $\phi$.

\section{Predicting Effect from Cause}

\begin{figure}[h]
\begin{center}
\includegraphics[width=8cm]{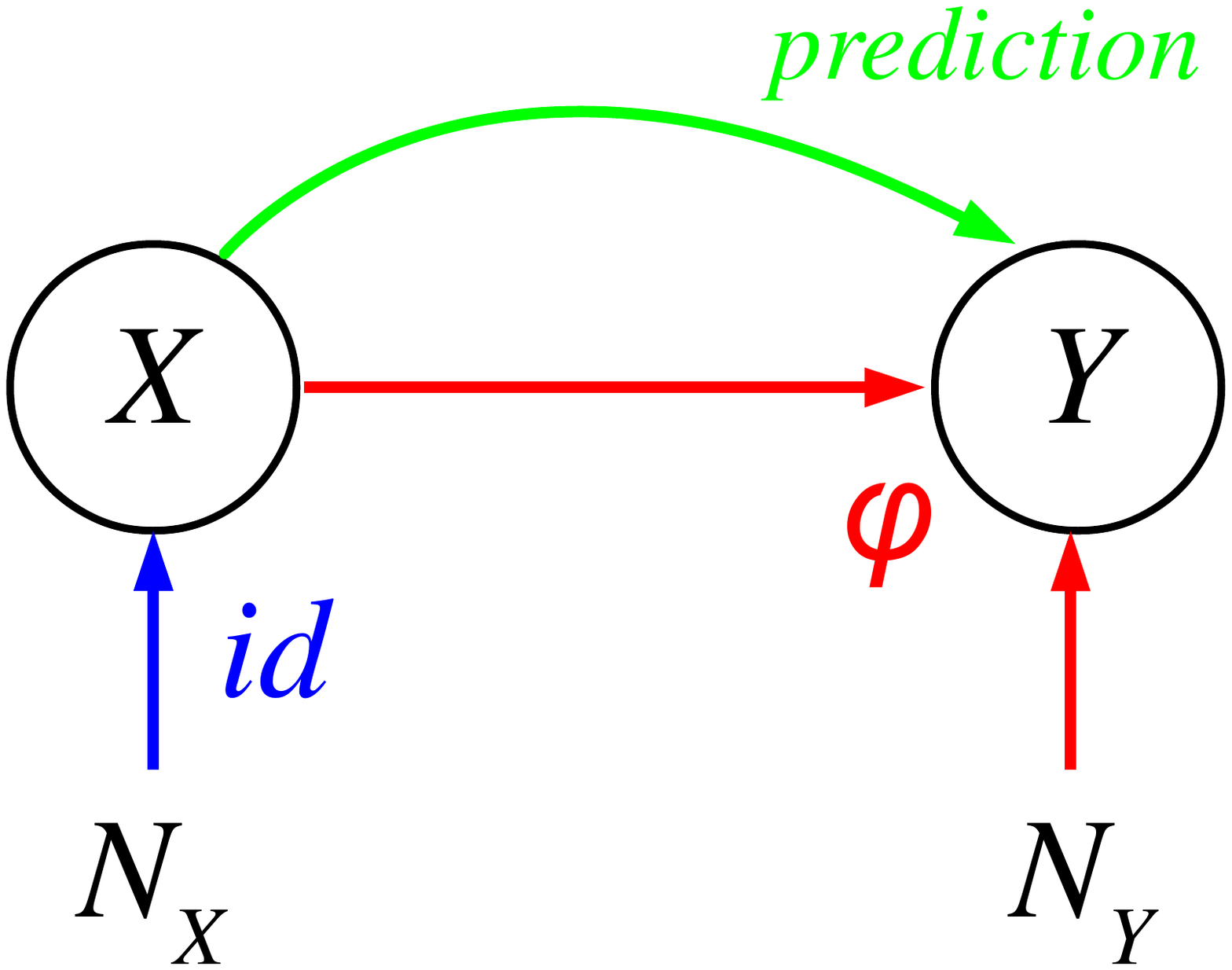}
\end{center}
\caption[font=small,labelsep=none]{\label{cause-effect-SEM-forward} Predicting effect $Y$ from cause $X$.}
\end{figure}

Let us consider the case where we are trying to estimate a function $f:\Xcal\to\Ycal$ or a conditional distribution $P(Y|X)$ in the causal direction, i.e., that $X$ is the cause and $Y$ the effect. Intuitively, this situation of {\em causal prediction} should be the 'easy' case since there exists a functional mechanism $\varphi$ which $f$ should try to mimic. We are interested in the question how robust (or invariant) the estimation is with respect to changes in the noise variables of the underlying functional causal model.

\subsection{Additional information about the input}

\subsubsection{Robustness w.r.t.\ input changes (distribution shift) \label{covshift}}

\paragraph{Given:} training points sampled from $P(X,Y)$ and an additional set of inputs sampled from $P'(X)$, with $P(X)\neq P'(X)$.

\paragraph{Goal:} estimate $P'(Y|X)$.

\paragraph{Assumption:} none.

\paragraph{Solution:} by independence of mechanism and input, there is no reason to assume that the observed change in $P(X)$ (i.e., in $P(N_X)$) entails a change in $P(Y|X)$, and we thus conclude $P'(Y|X)=P(Y|X)$. This scenario is referred to as \emph{covariate shift} \cite{SugKaw12}.

\subsubsection{Semi-supervised learning}

\paragraph{Given:} training points sampled from $P(X,Y)$ and an additional set of inputs sampled from $P(X)$.

\paragraph{Goal:} estimate $P(Y|X)$.

\paragraph{Note:} by independence of the mechanism, $P(X)$ contains no information about $P(Y|X)$. A more accurate estimate of $P(Y|X)$, as may be possible by the addition of the test inputs $P(X)$, does thus not influence an estimate of $P(Y|X)$, and semi-supervised learning (SSL) is pointless for the scenario in \fig{cause-effect-SEM-forward}.

\subsection{Additional information about the output}

\subsubsection{Robustness w.r.t.\ output changes}

\paragraph{Given:} training points sampled from $P(X,Y)$ and an additional set of outputs sampled from $P'(Y)$, with $P'(Y)\neq P(Y)$.

\paragraph{Goal:} estimate $P'(Y|X)$.

\paragraph{Assumption:} various options, e.g., an additive Gaussian noise model where $P(\phi(X))$ is indecomposable and $P'(\phi(X))$ is also indecomposable, if it is different from $P(\phi(X))$.

\paragraph{Solution:} first we need to decide whether 
$P(X)$ or $P(Y|X)$ has changed. This can be done using 
the method {\tt Localizing Distribution Change} (Subsection~\ref{localizing})
under appropriate assumptions (see above).
If $P(X)$ has changed, proceed as in Subsubsection~\ref{covshift}. 
If $P(Y|X)$ has changed, we can, estimate
$P'(Y|X)$ via {\tt Estimating Causal Conditional} (Subsection~\ref{estimating}).
Here, additive noise is  a sufficient assumption.

\subsubsection{Semi-supervised learning}

\label{SSLoutput} 

\paragraph{Given:} training points sampled from $P(X,Y)$ and an additional set of outputs sampled from $P(Y)$.

\paragraph{Goal:} estimate $P(Y|X)$.

\paragraph{Assumption:} $P(X,Y)$ has an additive noise model from $X$ to $Y$ and $P(Y)$ has a unique decomposition as convolution of two distributions, say $P(Y)=Q*R$.
This is, for instance satisfied if the noise is Gaussian and $P(\phi(C))$ is indecomposable.

\paragraph{Solution:} The additional outputs help because the decomposition tells us that either $P(N_Y)=Q$ or $P(N_Y)=R$.
The additive noise model learned from the $x,y$-pairs will probably tell us which of the alternatives is true. Knowing $P(N_E)$, the conditional $P(Y|X)$ reduces to learning $\phi$
from the $x,y$-pairs, which is certainly a weaker problem than learning $P(Y|X)$ would be in general.

\subsection{Additional information about input and output} 

\subsubsection{Transfer learning (only nosie changes)\label{transfer}}

\paragraph{Given:} training points sampled from $P(X,Y)$ and an additional set of points sampled from $P'(X,Y)$, with $P'(X,Y)\neq P(X,Y)$.

\paragraph{Goal:} estimate $P'(Y|X)$.

\paragraph{Assumption:} Additive noise where $\phi$ is invariant, but the noises can change.

\paragraph{Solution:} run {\tt conditional ANM} to output a single function, only enforcing independence of residuals separately for the two data sets (\sec{conditional-ANM}).

There is also a semi-supervised learning variant of this scenario: Given given a training set plus two unpaired sets from the two original marginals, then the extra sets help to better estimate $P(X,Y)$ because we have  argued in Subsubsection~\ref{SSLoutput} that additional $y$-values sampled from $P(Y)$ already help.

\subsubsection{Concept drift (only meachnism changes)}

\paragraph{Given:} training points sampled from $P(X,Y)$ and an additional set of points sampled from $P'(X,Y)$, with $P'(X,Y)\neq P(X,Y)$.

\paragraph{Goal:} estimate $P'(Y|X)$.

\paragraph{Assumption:} $N_X, N_Y$ invariant, but $\phi$ has changed.

\paragraph{Solution:} Apply ANM to points sampled from $P'(X,Y)$ to obtain $\phi$. Then $P'(Y|X)$ is given by 
\[
P'(Y|X)=P_{N_Y}(Y-\phi(X))\,.
\]

\section{Predicting Cause from Effect}
We now turn to the opposite direction, where we consider the effect as observed and we try to predict the value of the cause variable that led to it. This situation of {\em anticausal prediction} may seem unnatural, but it is actually ubiquitous in machine learning. Consider, for instance, the task of predicting the class label of a handwritten digit from its image. The underlying causal structure is as follows: a person intends to write the digit 7, say, and this intention causes a motor pattern producing an image of the digit 7 --- in that sense, it is justified to consider the class label $Y$ the cause of the image $X$.

\begin{figure}[h]
\begin{center}
\includegraphics[width=8cm]{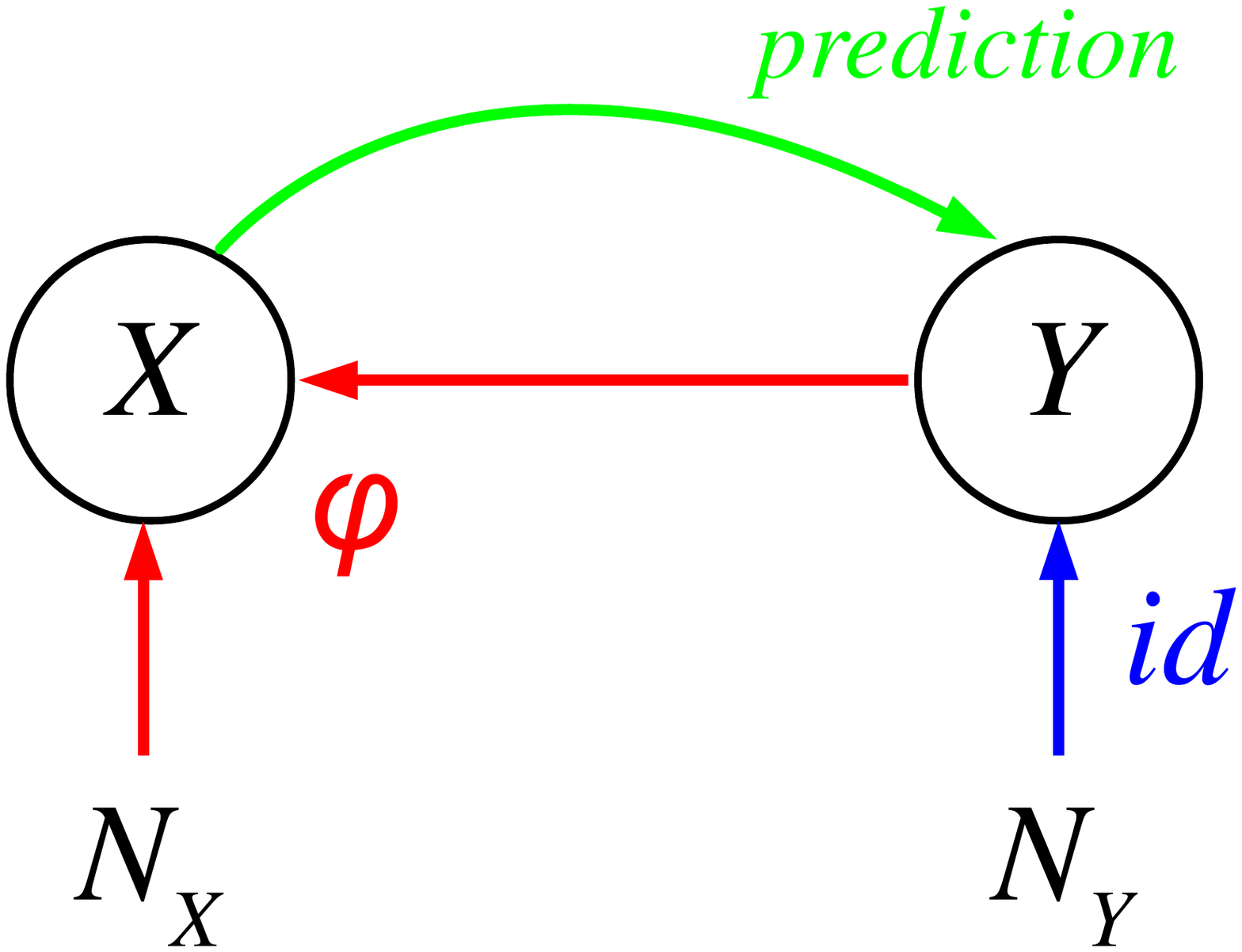}
\end{center}
\caption[font=small,labelsep=none]{\label{cause-effect-SEM-backward}Predicting cause $Y$ from effect $X$.}
\end{figure}

\subsection{Additional information about the input}

\subsubsection{Robustness w.r.t.\ input changes (distribution shift)\label{covshift-backwards}}

\paragraph{Given:} training points sampled from $P(X,Y)$ and an additional set of inputs sampled from $P'(X)$, with $P'(X)\neq P(X)$.\footnote{A related scenario is that we do not have additional data from $P'(X)$, but we want to still use our knowledge of the causal direction to learn a model that is somewhat robust w.r.t.\ changes of $P(X)$ due to changes in either $P(Y)$ or $P(X|Y)$.}

\paragraph{Goal:} estimate $P'(Y|X)$.

\paragraph{Assumption:} additive Gaussian noise with invertible function $\phi$ and indecomposable $P(\phi(Y))$ is sufficient.
Other assumptions are also possible, but invertibility of 
 the causal conditional $P(X|Y)$ is necessary in any case.

\paragraph{Solution:} 
We  apply {\tt Localizing Distribution Change} (Subsection~\ref{localizing})
to decide if $P(Y)$ or $P(X|Y)$ has changed.
In the first case, we can estimate $P'(Y)$ via {\tt Inverting Conditionals} (Subsection~\ref{inverting})
if we assume
that $P(X|Y)$ is 
an injective conditional.\footnote{This term will be introduced in Subsection~\ref{inverting}. Injectivity means that the input distribution can uniquely be computed from the output distribution. 
We will give examples of injective conditionals later.} 
From this we get $P'(X,Y)$, and then $$P'(Y|X)=\frac{P'(X,Y)}{\int P'(X,Y) dY}.$$

If, of the other hand, $P(X|Y)$ has changed, we can estimate $P'(X|Y)$ 
via {\tt Estimating Causal Conditionals} (Subsection~\ref{estimating}).

\subsubsection{Semi-supervised learning}

\paragraph{Given:} training points sampled from $P(X,Y)$ and an additional set of inputs sampled from $P(X)$.

\paragraph{Goal:} estimate $P(Y|X)$.

\paragraph{Assumption:} unclear. 

\paragraph{Note:} by dependence of the mechanism, $P(X)$ contains information about $P(Y|X)$. The additional inputs thus may allow a more accurate estimate of $P(X)$.\footnote{Note that a weak form of SSL could roughly work as followst: after learning a generative model for $P(X,Y)$ from the first part of the sample, we can use the additional samples from $P(X)$ to double check whether our model generates the right distribution for $P(X)$.}

Known methods for semi-supervised learning can indeed be viewed in this way. For instance, the cluster assumption says that points that lie in the same cluster of $P(X)$ should have the same $Y$; and the low density separation assumption says that the decision boundary of a classifier (i.e., the point where $P(Y|X)$ crosses $0.5$) should lie in a region where $P(Y)$ is small. The semi-supervised smoothness assumption says that the estimated function (which we may think of as the expectation of $P(Y|X)$ should be smooth in regions where $P(X)$ is large (for an overview of the common assumptions, see \cite{ChaSchZie06}). Some algorithms assume a model for the causal mechanism, $P(X|Y)$, which is usually a Gaussian distribution or mixture of Gaussians, and learn it on both labeled and unlabeled data~\cite{Zhu09}.  Note that all these assumptions translate properties of $P(X)$ into properties of $P(Y|X)$.

Using a more accurate estimate of $P(X)$, we could also try to proceed as in Subsubsection~\ref{covshift-backwards}.\footnote{However, in this case we do not have the two alternatives of whether $P(Y)$ or $P(X|Y)$ has changed. The question now should be: given a better estimate of $P(X)$, does that change our estimate of $P(Y)$, or of $P(X|Y)$?}

\subsection{Additional information about the output}

\subsubsection{Robustness w.r.t.\ output changes}

\paragraph{Given:} training points sampled from $P(X,Y)$ and an additional set of outputs sampled from $P'(Y)$, with $P'(Y)\neq P(Y)$.

\paragraph{Goal:} estimate $P'(Y|X)$.

\paragraph{Assumption:} none.

\paragraph{Solution:} independence of mechanism implies $P'(X|Y)=P(X|Y)$, hence $P'(X,Y) = P(X|Y) P'(Y)$. From this, we compute 
$$P'(Y|X) = \frac{P'(X|Y) P'(Y)}{\int P'(X,Y)  dY}.$$

There may also be room for a semi-supervised learning variant: suppose we have additional output observations rather than additional inputs as in standard SSL --- in which situations does this help?

\subsection{Additional information about input and output}

\subsubsection{Robustness w.r.t.\ changes of input and output noise (transfer learning)}

\paragraph{Given:} training points sampled from $P(X,Y)$ and an additional set of points sampled from $P'(X,Y)$, with $P'(X,Y)\neq P(X,Y)$.

\paragraph{Goal:} estimate $P'(Y|X)$.

\paragraph{Assumption:} additive noise where $\phi$ is invariant, but the noises can change.

\paragraph{Solution:} analogous to Subsection~\ref{transfer}, but use the model backwards in the end.

\subsubsection{Concept drift (changes of the mechanism)}

\paragraph{Given:} training points sampled from $P(X,Y)$ and an additional set of points sampled from $P'(X,Y)$, with $P'(X,Y)\neq P(X,Y)$.

\paragraph{Goal:} estimate $P'(Y|X)$.

\paragraph{Assumption:} $N_X, N_Y$ invariant, but $\phi$ has changed to $\phi'$.

\paragraph{Solution:}
We can learn $\phi'$ from $P'(X,Y)$ and 
then estimate the entire distribution $P'(X,Y)$ using the estimations of our distributions $P(N_X)$ and $P(N_Y)$ obtained from observing those $x,y$ pairs that were taken from $P(X,Y)$.

\section{Modules}

\subsection{Inverting Conditionals\label{inverting}}

We can think of a conditional $P(Y|X)$ as a mechanism that transforms $P(X)$ into $P(Y)$. In some cases, we do not loose any information by this mechanism:

\begin{Def}[injective conditionals]
a conditional distribution $P(Y|X)$ is called injective if there  are no two
distributions $P(X)\neq P'(X)$ such that
\[
\int P(y|x) P(x) dx = \int P(y|x) P'(x) dx\,.
\]
\end{Def}

\begin{Ex}[full rank stochastic matrix]
Let $X,Y$ have finite range. Then $P(Y|X)$ is given by a stochastic matrix $M$ and
is injective if and only if $M$ has full rank. Note that this is only possible if
$|\Xcal| \leq |\Ycal|$.
\end{Ex}

\begin{Ex}[Post-nonlinear model]
 Let $X,Y$ be real-valued and 
\[
Y=\psi (\phi(X)+N_Y) \hbox{ with } N_Y \independent X\,,
\]
be a post-nonlinear model where $\phi$ and $\psi$ are injective.
Then the distribution of $Y$ uniquely determines the distribution of
$\phi(X)+N_Y$ because $\psi$ is invertible. This in turn, uniquely determines the distribution of $\phi(X)$ provided that the convolution with $P(N_Y)$ is invertible.
Since $\psi$ is invertible, this determines the distribution of $X$ uniquely.
\end{Ex}

Note that additive noise models with injective $\phi$ are a special case of a post-non-linear model by setting $\psi:=id$.

\subsection{Localizing distribution change\label{localizing}}

Given data points sampled from $P(C,E)$ and additional points from $P'(E)\neq P(E)$, we wish to decide whether $P(C)$ or $P(E|C)$ has changed.
Assume 
\begin{eqnarray*}
E=\phi(C)+N_E\,,
\end{eqnarray*}
with the same $\phi$ for both distributions $P(E,C)$ and $P'(E,C)$, but the distribution of the noise
$N_E$ or  the distribution of $C$ changes. 
Let $P(\phi(C))$ denote the distribution of $\phi(C)$.\footnote{Explicitly, it is derived from the distribution of $C$ by $P(\phi(C)\in A)=P(C\in \phi^{-1}(A))$.}

 Then
the distributions of the effect
are given by
\begin{eqnarray*}
P(E)&=&P(\phi(C))*P(N_E)\\
P'(E)&=&P'(\phi(C))*P'(N_E)\,,
\end{eqnarray*}
where either $P'(\phi(C))=P(\phi(C))$ or $P'(N_E)=P(N_E)$.
To decide which of these cases is true, we
first estimate $\phi$ from the first data set, and then 
apply a deconvolution with $P(\phi(C))$ (denoted with $P(\phi (C)) *^{-1} \cdot$)  
or $P(N_E)$
to $P'(E)$ and check whether (1) $ P(\phi (C))*^{-1} P'(E) $ or (2) $ P(\phi (C)) *^{-1}  P'(E)$ is
a probability distribution. 
Below we will dicuss one possible set of assumptions that ensure that exactly one of the alternatives should is true. 
In case (1), $P(E|C)$ has changed. In case (2), $P(C)$ has changed.

To show that there are (not too artificial) asumptions that render the problem solvable, assume that 
$P(\phi(C))$ and $P'(\phi(C))$ are indecomposable and $P(N_E)$ and $P'(N_E)$ are Gaussian with zero mean. 
Then the distribution $P(E)=P(\phi(C))*P(N_E)$ uniquely determines $P(\phi(C))$ by deconvolving $P(E)$ with the Gaussian 
of maximal possible width that still yields a probability distribution. 

We are aware that there exist situations where both cases are possible. For instance, consider the example in which $P(f(C))$ follows a uniform distribution, $P(N_E) \sim \mathcal{N}(0,1)$, while when generating $P'(E)$, $P'(f(C)) = P(f(C))$ and $P'(N_E)\sim \mathcal{N}(0,2)$.  That is, when generating the new data, only $P(E|C)$ was changed.  However, applying the deconvolution with $P(N_E)$ to $P'(E)$ results in $P'(E) *^{-1} P(N_E) = P(f(C)) * \big(P'(N_E) *^{-1} P(N_E) \big) = P(f(C)) * \mathcal{N}(0, 2-1) = P(f(C)) * \mathcal{N}(0, 1)$, which still corresponds a valid distribution. Consequently, we have to conclude that both cases are possible.

Despite the examples where the proposed method fails, the proposed method also works in -- hopefully -- many situations.  For instance, now let us switch the roles of $P(E)$ and $P'(E)$ in the example above, or in other words, suppose $P(N_E)\sim \mathcal{N}(0,2)$ and $P'(N_E)\sim \mathcal{N}(0,1)$. In this example deconvolving $P'(E)$ with $P(N_E)$ gives $P'(E) *^{-1} P(N_E) = P(f(C)) * P'(N_E) *^{-1} P(N_E)  = P(f(C)) *^{-1} \mathcal{N}(0, 1)$, which is not a valid distribution. That is, in this example we can make the decision that $P(E|C)$ has changed. We are working on the conditions to guarantee that only one of the two cases is possible.

\subsection{Estimating causal conditionals\label{estimating}}

Given $P'(E)$, estimate $P'(E|C)$ under the assumption that $P(C)$ 
remained constant. 
Assume that $P(E,C)$ and $P'(E,C)$ have been generated by the additive noise model
\[
E=\phi(C)+N_E\,,
\]
with the same $P(C)$ and $f$, while the distribution of $N_E$ has changed.
We have
\begin{eqnarray*}
P(E)&=&P(\phi(C))*P(N_E)\,,\\
P'(E)&=&P(\phi(C))*P'(N_E)\,.
\end{eqnarray*}
Hence, $P'(N_E)$ can be obtained by the deconvolution
\[
P'(N_E)= P(\phi(C)) *^{-1} P'(E) \,. 
\]
This way, we can compute the new conditional $P'(E|C)$.

\subsection{Conditional {\tt ANM}\label{conditional-ANM}}
Given two data sets generated by
\begin{equation}\label{eq:ANM1}
E = \phi(C) + N_E
\end{equation}
and
\begin{equation}\label{eq:ANM2}
E' = \phi(C') + N_E',
\end{equation}
respectively. We apply the algorithm of \cite{MooJanPetSch09} to obtain the shared function $\phi$, enforcing separate independence $C\independent N_E$ and $C'\independent N_E'$. 

This can be interpreted as a {\tt ANM} model enforcing conditional independence in
\begin{equation}\label{eq:ANM3}
E|i = \phi(C|i) + N_E|i,
\end{equation}
where $i\in\{1,2\}$ is an index, and $C|i \independent N_E|i$.

\paragraph{Acknowledgement}
We thank Joris Mooij, Bob Williamson, Vladimir Vapnik, Jakob Zscheischler and Eleni Sgouritsa for helpful discussions.

\bibliography{bibfile2}

\end{document}